\documentclass{article}
\usepackage{nips15submit_e,times}
\nipsfinalcopy

\usepackage[numbers]{natbib}
\usepackage{amssymb}
\usepackage{amsmath}
\usepackage{graphicx}
\usepackage{color}
\usepackage{psfrag}
\usepackage{stfloats}
\usepackage{multirow}
\usepackage{rotating}
\usepackage{amssymb}
\usepackage{paralist}
\usepackage{subfigure} 
\usepackage{wrapfig}
\usepackage{booktabs}
\usepackage{sidecap}

\usepackage{url}

\newcommand{\bmx}[0]{\begin{bmatrix}}
\newcommand{\emx}[0]{\end{bmatrix}}

\newenvironment{bookman}{\scriptsize\fontfamily{bch}\selectfont}{\par}

\begin{document}

\title{Bidirectional Recurrent Neural Networks as Generative Models}

\author{Mathias Berglund \\ Aalto University, Finland \And Tapani Raiko \\ Aalto University, Finland \And Mikko 
Honkala  \\ Nokia Labs, Finland  \And Leo K\"{a}rkk\"{a}inen \\ Nokia Labs, Finland  \And Akos Vetek \\ Nokia Labs, Finland 
\And 
Juha Karhunen \\ Aalto University, Finland }


\maketitle

\begin{abstract}

Bidirectional recurrent neural networks (RNN) are trained to predict both in the 
positive and negative time directions simultaneously. They have not been 
used commonly in unsupervised tasks, because a probabilistic interpretation 
of the model has been difficult. 
Recently, two different frameworks, GSN and NADE, provide a connection between reconstruction and probabilistic modeling, which makes the interpretation possible. 
As far as we know, neither GSN or NADE have been studied in the context of time series before.
As an example of an unsupervised task, we study 
the problem of filling in gaps in high-dimensional time series with complex dynamics. 
Although unidirectional RNNs have recently been trained 
successfully to model such time series, inference in the negative time 
direction is non-trivial. 
We propose two probabilistic interpretations of bidirectional RNNs that can be 
used to reconstruct missing gaps efficiently. Our experiments on text data show 
that both proposed methods are much more accurate than unidirectional 
reconstructions, although a bit less accurate than a computationally complex 
bidirectional Bayesian inference on the unidirectional RNN. We also provide 
results on music data for which the Bayesian inference is computationally infeasible, demonstrating the scalability of the proposed methods.


\end{abstract}

\section{Introduction}

Recurrent neural networks (RNN) have recently been trained successfully for 
time series modeling, and have been used to achieve state-of-the-art results in 
supervised tasks including handwriting recognition \citep{graves2009novel} and 
speech recognition \citep{graves2013speech}. RNNs have also been used 
successfully in unsupervised learning of time series 
\citep{sutskever2011generating,boulanger2012modeling}. 

Recently, RNNs have also been used to generate sequential data
\citep{bahdanau2014neural} in a machine translation context, which further
emphasizes the unsupervised setting. \citet{bahdanau2014neural}
used a bidirectional RNN to encode a phrase into a vector, but settled for
a unidirectional RNN to decode it into a translated phrase, perhaps because
bidirectional RNNs have not been studied much as generative models.
Even more recently, \citet{maas2014first} used a deep bidirectional RNN in speech
recognition, generating text as output.

Missing value reconstruction is interesting in at least three different senses. 
Firstly, it can be used to cope with data that really has missing values. 
Secondly, reconstruction performance of artificially missing values can be used 
as a measure of performance in unsupervised learning \citep{raiko2001missing}. 
Thirdly, reconstruction of artificially missing values can be used as a 
training criterion \citep{brakel2013training,goodfellow2013multi,uria2014deep}. 
While traditional RNN training criterions correspond to one-step prediction, training to 
reconstruct longer gaps can push the model towards concentrating on longer-term 
predictions. Note that the one-step prediction criterion is typically used even in approaches
that otherwise concentrate on modelling long-term dependencies 
\citep[see e.g.][]{mikolov2014learning,koutnik2014clockwork}.

When using unidirectional RNNs as generative models, it is straightforward to 
draw samples 
from the model in sequential order. However, inference is not trivial in 
smoothing tasks, where we want to evaluate probabilities for missing values in 
the middle of a time series. For discrete data, inference with gap sizes of one is 
feasible - however, inference with larger gap sizes becomes exponentially more 
expensive. Even sampling can be exponentially expensive with respect to the gap 
size.

One strategy used for training models that are used for filling in gaps is to 
explicitly train the model with missing data \citep[see e.g.][]{brakel2013training}.
However, such a criterion has not to our knowledge 
yet been used and thoroughly evaluated compared with other inference strategies 
for RNNs.

In this paper, we compare different methods of using RNNs to infer missing values 
for binary time series data. We evaluate the performance of two generative models that
rely on bidirectional RNNs, and compare them to inference using a unidirectional RNN.
The proposed methods are very favourable in terms of scalability.

\section{Recurrent Neural Networks}
\label{sec:rnns}

Recurrent neural networks \citep{rumelhart1986learning,haykin2009neural} can be seen as extensions of the standard feedforward 
multilayer perceptron networks, where the inputs and outputs are sequences 
instead of individual observations.

Let us denote the input to a recurrent neural network by $\mathbf{X} = 
\left\lbrace \mathbf{x}_t \right\rbrace$ where $\mathbf{x}_t \in \mathbb{R}^N$ 
is an input vector for each time step $t$. Let us further denote the output as 
$\mathbf{Y} = \left\lbrace \mathbf{y}_t \right\rbrace$ where $\mathbf{y}_t \in 
\mathbb{R}^M$ is an output vector for each time step $t$. Our goal is to model 
the distribution $P \left( \mathbf{Y} | \mathbf{X} \right)$. Although RNNs 
map input sequences to output sequences, we can use them in an unsupervised manner by letting 
the RNN predict the next input. We can do so by setting $\mathbf{Y} = 
\left\lbrace \mathbf{y}_t = \mathbf{x}_{t+1} \right\rbrace$.

\subsection{Unidirectional Recurrent Neural Networks}

\begin{figure}[t]
    \centering
    \includegraphics[width=0.45\textwidth]{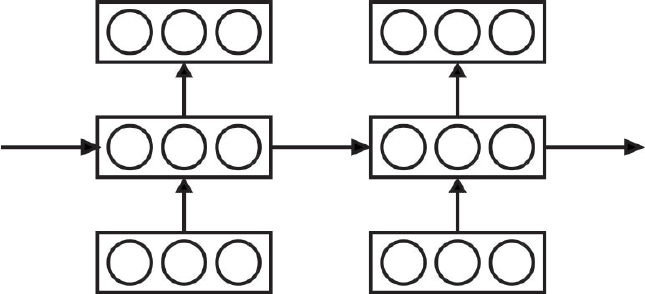}
    \includegraphics[width=0.45\textwidth]{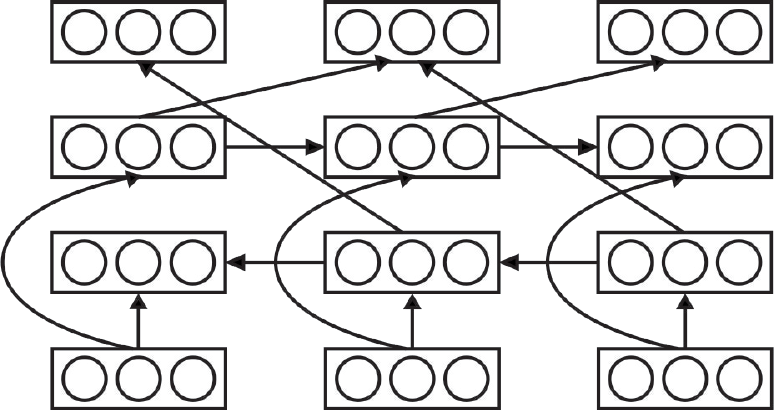}
    \caption{Structure of the simple RNN (left) and the bidirectional RNN (right).}
    \label{fig:rnn_architecture}
    \vspace{-3mm}
\end{figure}

The structure of a basic RNN with one hidden layer is  illustrated in 
Figure~\ref{fig:rnn_architecture}, where the output $\mathbf{y}_t$ is determined 
by

\begin{equation}
P\left(\mathbf{y}_t \mid \{\mathbf{x}_d\}_{d=1}^t\right) = \mathrm{\phi} \left( 
\mathbf{W}_\mathrm{y} \mathbf{h}_t + \mathbf{b}_\mathrm{y} \right)
\label{eq:uni_y}
\end{equation}
where
\begin{equation}
\mathbf{h}_t = \tanh \left( \mathbf{W}_\mathrm{h} \mathbf{h}_{t-1} + \mathbf{W}_\mathrm{x} \mathbf{x}_{t} + \mathbf{b}_\mathrm{h} \right)
\label{eq:uni_h}
\end{equation}
and $\mathbf{W}_\mathrm{y}$, $\mathbf{W}_\mathrm{h}$, and 
$\mathbf{W}_\mathrm{x}$ are the weight matrices connecting the hidden to output 
layer, hidden to hidden layer, and input to hidden layer, respectively. 
$\mathbf{b}_\mathrm{y}$ and $\mathbf{b}_\mathrm{h}$ are the output and hidden 
layer bias vectors, respectively. 
Typical options for the 
final nonlinearity $\mathrm{\phi}$ are the softmax function for classification or categorical prediction
tasks, or independent Bernoulli variables with sigmoid functions for other 
binary prediction tasks. In this form, 
the RNN therefore evaluates the output $\mathbf{y}_t$ based on information 
propagated through the hidden layer that directly or indirectly depends on the 
observations $\{\mathbf{x}_d\}_{d=1}^t= \{\mathbf{x}_1, \dots, 
\mathbf{x}_t\}$. 

\subsection{Bidirectional Recurrent Neural Networks}
\label{sec:bidirectional}

Bidirectional RNNs 
(BRNN) \citep{schuster1997bidirectional,baldi1999exploiting} 
extend the unidirectional RNN by introducing a second hidden layer, where the 
hidden to hidden connections flow in opposite temporal order. The model is 
therefore able to exploit information both from the past and the future. 

The output $\mathbf{y}_t$ is traditionally determined by
\begin{equation*}
P\left(\mathbf{y}_t \mid \{\mathbf{x}_d\}_{d\neq t}\right) = \mathrm{\phi} 
\left( \mathbf{W}_\mathrm{y}^\mathrm{f} 
\mathbf{h}_{t}^\mathrm{f} + \mathbf{W}_\mathrm{y}^\mathrm{b} 
\mathbf{h}_{t}^\mathrm{b} + \mathbf{b}_y \right), 
\end{equation*}
but we propose the use of
\begin{equation}
P\left(\mathbf{y}_t \mid \{\mathbf{x}_d\}_{d\neq t}\right) = \mathrm{\phi} 
\left( \mathbf{W}_\mathrm{y}^\mathrm{f} 
\mathbf{h}_{t-1}^\mathrm{f} + \mathbf{W}_\mathrm{y}^\mathrm{b} 
\mathbf{h}_{t+1}^\mathrm{b} + \mathbf{b}_y \right)
\label{eq:bi_y}
\end{equation}
where
\begin{align}
\mathbf{h}_t^\mathrm{f} &= \tanh \left( \mathbf{W}_\mathrm{h}^\mathrm{f} 
\mathbf{h}^\mathrm{f} _{t-1} + \mathbf{W}_\mathrm{x}^\mathrm{f} \mathbf{x}_{t} + 
\mathbf{b}_\mathrm{h}^\mathrm{f} \right)
\label{eq:bi_hf}
\\
\mathbf{h}_t^\mathrm{b} &= \tanh \left( \mathbf{W}_\mathrm{h}^\mathrm{b} 
\mathbf{h}^\mathrm{b}_{t+1} + \mathbf{W}_\mathrm{x}^\mathrm{b} \mathbf{x}_{t} 
+ \mathbf{b}_\mathrm{h}^\mathrm{b} \right).
\label{eq:bi_hb}
\end{align}

The structure of the BRNN is illustrated in Figure~\ref{fig:rnn_architecture} (right). 
Compared with the regular RNN, the forward and backward directions have separate
non-tied weights and hidden activations, and are denoted by the superscript f and b for 
forward and backward, respectively. 
Note that the connections are acyclic. 
Note also that in the proposed formulation, $\mathbf{y}_{t}$ does not get 
information from $\mathbf{x}_{t}$. 
We can therefore use the model in an unsupervised 
manner to predict one time step given all other time steps in the input 
sequence simply by setting $\mathbf{Y}=\mathbf{X}$. 

\section{Probabilistic Interpretation for Unsupervised Modelling}

Probabilistic unsupervised modeling for sequences using a unidirectional RNN is 
straightforward, as the joint distribution for the whole sequence is simply the 
product of the individual predictions:
\begin{align}
P_{\mathrm{unidirectional}}(\mathbf{X}) = \prod_{t=1}^T P(\mathbf{x}_t \mid \{\mathbf{x}_d\}_{d=1}^{t-1}). 
\end{align}

For the BRNN, the situation is more complicated. The network gives 
predictions for individual outputs given all the others, and the joint 
distribution cannot be written as their product. We propose two solutions for 
this, denoted by \emph{GSN} and \emph{NADE}.

{\bf GSN}
Generative Stochastic Networks (GSN) \citep{bengio2013generalized} use a denoising 
auto-encoder to estimate the data distribution
as the asymptotic distribution of the Markov chain that alternates between 
corruption and denoising.
The resulting distribution is thus defined only implicitly, and cannot be written analytically.
We can define a corruption function that masks $\mathbf{x}_{t}$ as missing, 
and a denoising function that reconstructs it from the others. It turns out 
that one feedforward pass of the BRNN does exactly that.

Our first probabilistic interpretation is thus that the joint distribution 
defined by a BRNN is the asymptotic distribution of a process that 
replaces one observation vector $\mathbf{x}_{t}$ at a time in $\mathbf{X}$ by 
sampling it from $P_{\mathrm{BRNN}}(\mathbf{x}_{t} \mid \{\mathbf{x}_d\}_{d\neq t})$. In 
practice, we will start from a random initialization and use Gibbs sampling.

{\bf NADE}
The Neural Autoregressive Distribution Estimator (NADE) 
\citep{uria2014deep} defines a probabilistic model by reconstructing missing 
components of a vector one at a time in a random order, starting from a fully 
unobserved vector. Each reconstruction is given by an auto-encoder network that 
takes as input the observations so far and an auxiliary mask vector that 
indicates which values are missing.

We extend the same idea for time series. Firstly, we concatenate an auxiliary 
binary element to input vectors to indicate a missing input. The joint 
distribution of the time series is defined by first drawing a random permutation 
$o_d$ of time indices $1\dots T$ and then setting data points observed one 
by one in that order, starting from a fully missing sequence: 
\begin{align}
P_{\mathrm{NADE}}(\mathbf{X}\mid o_d) = \prod_{d=1}^T P(\mathbf{x}_{o_d} \mid \{\mathbf{x}_{o_{e}}\}_{e=1}^{d-1}).
\end{align}
In practice, the BRNN will be trained with some inputs marked as missing, while 
all the outputs are observed. See Section \ref{sec:categ_details} for more training details.


\section{Filling in gaps with Recurrent Neural Networks}

The task we aim to solve is to fill in gaps of multiple consecutive data points 
in high-dimensional binary time series data. The inference is not trivial for 
two reasons: firstly, we reconstruct multiple consecutive data points, which are likely 
to depend on each other, and secondly, we fill in data in the middle of a time 
series and hence need to consider the data both before and after the gap.

For filling in gaps with the GSN approach, we first train a bidirectional RNN 
to estimate $P_{\mathrm{BRNN}}(\mathbf{x}_{t} \mid \{\mathbf{x}_d\}_{d\neq t})$. In order to 
achieve that, we use the structure presented in Section~\ref{sec:bidirectional}. 
At test time, the gap is first initialized to random values, after which the 
missing values are sampled from the distribution $P_{\mathrm{BRNN}}(\mathbf{x}_{t} \mid 
\{\mathbf{x}_d\}_{d\neq t})$ one by one in a random order repeatedly to approximate the stationary distribution. For 
the RNN structures used in this paper, the computational complexity of this 
approach at test time is $\mathcal{O}((dc+c^2)(T+gM))$ where $d$ is the 
dimensionality of a data point, $c$ is the number of hidden units in the RNN, 
$T$ is the number of time steps in the data, $g$ is the length of the gap and 
$M$ is the number of Markov chain Monte Carlo (MCMC) steps used for inference.

For filling in gaps with the NADE approach, we first train a bidirectional RNN 
where some of the inputs are set to a separate missing value token. At test 
time, all data points in the gap are first initialized with this token, after 
which each missing data point is reconstructed once until the whole gap is 
filled. Computationally, the main difference to GSN is that we do not have to 
sample each reconstructed data point multiple times, but the reconstruction is done in as many 
steps as there are missing data points in the gap. For the RNN structures used 
in this paper, the computational complexity of this approach at test time is 
$\mathcal{O}((dc+c^2)(T+g))$ where $d$ is the dimensionality of a data point, 
$c$ is the number of hidden units in the RNN, $g$ is the length of the gap and 
$T$ is the number of time steps in the data.

In addition to the two proposed methods, one can use a unidirectional RNN to 
solve the same task. We call this method \emph{Bayesian MCMC}. Using a 
unidirectional RNN for the task of filling in gaps is not trivial, as we need to take into account 
the probabilities of the values after the gap, which the model does not 
explicitly do. We therefore resort to a similar approach as the GSN approach, 
where we replace the $P_{\mathrm{BRNN}}(\mathbf{x}_{t} \mid \{\mathbf{x}_d\}_{d\neq t})$ with a 
unidirectional equivalent for the Gibbs sampling. As the unidirectional RNN 
models conditional probabilities of the form $P_{\mathrm{RNN}}(\mathbf{x}_t \mid 
\{\mathbf{x}_d\}_{d=1}^{t-1})$, we can use Bayes' theorem to derive:

\begin{align}
\hspace{2em}&\hspace{-2em} P_{\mathrm{RNN}}\left( \mathbf{x}_{t} = \mathbf{a} \mid \{\mathbf{x}_{d}\}_{d\neq t} \right) 
\\ 
& \propto P_{\mathrm{RNN}} \left( \mathbf{x}_t = \mathbf{a} \mid \{\mathbf{x}_d\}_{d=1}^{t-1} 
\right) P_{\mathrm{RNN}} \left(\{\mathbf{x}_e\}_{e=t+1}^{T} \mid \mathbf{x}_t = 
\mathbf{a}, \{\mathbf{x}_d\}_{d=1}^{t-1} \right) \\
&= \prod_{\tau=t}^T P_{\mathrm{RNN}}(\mathbf{x}_{\tau} \mid \{\mathbf{x}_d\}_{d=1}^{\tau-1}) \Bigr|_{\mathbf{x}_t=\mathbf{a}}
\label{eq:onewayrnn}
\end{align}
where $P_{\mathrm{RNN}}(\mathbf{x}_{\tau} \mid \{\mathbf{x}_d\}_{d=1}^{\tau-1})$ is directly 
the output of the unidirectional RNN given an input sequence $\mathbf{X}$, 
where one time step $t$, i.e. the one we Gibbs sample, is replaced by a proposal 
$\mathbf{a}$. The problem is that we have to go through all possible proposals 
$\mathbf{a}$ separately to evaluate the probability $P\left( \mathbf{x}_{t} = 
\mathbf{a} | \{\mathbf{x}_d\}_{d\neq t} \right)$. We therefore have to evaluate 
the product of the outputs of the unidirectional RNN for time steps $t \ldots 
T$ for each possible $\mathbf{a}$.

In some cases this is feasible to evaluate. For categorical data, 
e.g. text, there are as many possible values for $\mathbf{a}$ as there are 
dimensions\footnote{For character-based text, the number of dimensions is the 
number of characters in the model alphabet.}. However, for other binary data the 
number of possibilities grows exponentially, and is clearly not feasible to 
evaluate. For the RNN structures used in this paper, the computational complexity 
of this approach at test time is $\mathcal{O}((dc+c^2)(T+aTM))$ where $a$ is the 
number of different values a data point can have, $d$ is the dimensionality of a 
data point, $c$ is the number of hidden units in the RNN, $T$ is the number of 
time steps in the data, and $M$ is the number of MCMC steps used for inference. 
The critical difference in complexity to the GSN approach is the coefficient 
$a$, that for categorical data takes the value $d$, for binary vectors $2^d$ and 
for continuous data is infinite.


As a simple baseline model, we also evaluate the \emph{one-gram} log-likelihood 
of the gaps. The one-gram model assumes a constant context-independent 
categorical distribution for the categorical 
task, or a vector of factorial binomial probabilities for the structured 
prediction task:
\begin{equation*}
P_{\mathrm{one-gram}}\left(\mathbf{y}_t \right) = \mathrm{f} \left( \mathbf{b}_y \right).
\end{equation*}
This can be done in $\mathcal{O}(dg)$.

We also compare to \emph{one-way inference}, where the data points in the gap are 
reconstructed in order without taking the future context into account, using Equations 
(\ref{eq:uni_y}) and (\ref{eq:uni_h}) directly.
The computational complexity is $\mathcal{O}((dc+c^2)T)$.

\section{Experiments}
\label{sec:exp}

We run two sets of experiments: one for a categorical prediction task, and one for a 
binary structured prediction task. In the categorical prediction task we fill in gaps of 
five characters in Wikipedia text, while in the structural prediction task we 
fill in gaps of five time steps in different polyphonic music data sets.

\subsection{Training details for categorical prediction task}
\label{sec:categ_details}

For the categorical prediction task, we test the performance of the two 
proposed methods, GSN and NADE. In addition, we compare the performance to MCMC using Bayesian
inference and one-way inference with a unidirectional RNN. We therefore have to train 
three different RNNs, one for each method.

Each RNN is trained as a predictor network, where the character at each 
step is predicted based on all the previous characters (in the case of the 
RNN) or all the previous and following characters (in the case 
of the BRNNs). We use the same data set as \citet{sutskever2011generating}, which consists of 2GB of English text from Wikipedia. For training, we follow
a similar strategy as \citet{NIPS2013_5166}.
The characters are encoded as one-hot binary vectors with a dimensionality of $d=96$ characters and the output is modelled 
with a softmax distribution. We 
train the unirectional RNN with string lengths of $T=250$ characters, where the 
error is propagated only from the last 200 outputs. In the BRNN we 
use string length of $T=300$ characters, where the error is propagated from the 
middle 200 outputs. We therefore avoid propagating the gradient from predictions 
that lack long temporal context.

For the BRNN used in the NADE method, we add one dimension to the one-hot input 
which corresponds to a missing value token. During training, in each minibatch 
we mark $g=5$ consecutive characters every 25 time steps as a gap. During 
training, the error is propagated only from these gaps. For each gap, we 
uniformly draw a value from 1 to 5, and set that many characters in the gap to the missing value token. The model is therefore trained to 
predict the output in different stages of inference, where a number of the 
inputs are still marked as missing. For comparison, we also train a similar network,
but without masking. In that variant, the error is therefore propagated from all time steps.
We refer to ``NADE'' masked and ``NADE no mask'', respectively, for these two training methods.

For all the models, the weight elements are drawn from the uniform 
distribution: $w_{i,j}\sim \mathcal{U}\left[-s, s\right]$ where $s=1$ for the 
input to hidden layer, and following \citet{glorot2010understanding}, where $s=\sqrt{6/\left(d_{in} + d_{out}\right)}$ for the 
hidden-to-hidden and the hidden-to output layers. The biases are initialized to 
zero.

We use $c=1000$ hidden units in the unidirectional RNN and $c=684$ hidden units in the 
two hidden layers in the BRNNs. The number of parameters in the two model types 
is therefore roughly the same. In the recurrent layers, we set the recurrent activation connected to the first time step to zero.

The networks are trained using stochastic gradient descent and the gradient is 
calculated using backpropagation through time. We use a minibatch size of 40, 
i.e. each minibatch consists of 40 randomly sampled sequences of length 250. As 
the gradients tend to occasionally ``blow up'' when training RNNs 
\citep{bengio1994learning,pascanu2013difficulty}, we normalize the gradients at 
each update to have length one. The step size is set to 0.25 for all layers in 
the beginning of training, and it is linearly decayed to zero during training. 
As training the model is very time-consuming\footnote{We used about 8 weeks of GPU time for the reported results.}, we do not optimize the 
hyperparameters, or repeat runs to get confidence intervals around the 
evaluated performances.

\subsection{Training Details for the Binary Structured Prediction Task}

In the other set of experiments, we use four polyphonic music data sets 
\citep{boulanger2012modeling}. The data sets consist of at least 7 hours of polyphonic 
music each, where each data point is a binary $d=88$-dimensional vector that represents one 
time step of MIDI-encoded music, indicating which of the 88 keys of a piano are pressed.
We test the performance of the two proposed 
methods, but omit training the unidirectional RNNs as the computational complexity of the Bayesian MCMC is prohibitive ($a=2^{88}$).

We train all models for $50~000$ updates in minibatches of $\approx 3~000$ 
individual data points\footnote{A minibatch can therefore consist of e.g. $100$ 
musical scores, each of length $T=30$.}. As the data sets are small, we select the 
initial learning rate on a grid of $\left\lbrace 0.0001, 0.0003, \ldots, 0.3, 
1\right\rbrace$ based on the lowest validation set cost. We use no ``burn-in'' as 
several of the scores are fairly short, and therefore do not specifically mask 
out values in the beginning or end of the data set as we did for the text data.

For the NADE method, we use an additional dimension as a missing value token in 
the data. For the missing values, we set the missing value token to one and the 
other dimensions to zero.

Other training details are similar to the categorical prediction task.

\subsection{Evaluation of Models}
At test time, we evaluate the models by calculating the mean log-likelihood of the 
correct value of gaps of five consecutive missing values in test data.

In the GSN and Bayesian MCMC approaches, we first set the five values in the gap to a random value 
for the categorical prediction task, or to zero for the structured prediction task. We 
then sample all five values in the gap in random order, and repeat the procedure 
for $M=100$ MCMC steps\footnote{$M=100$ MCMC steps means that each value in the gap of $g=5$ will be resampled $M/g=20$ times}. For 
evaluating the log-likelihood of the correct value for the string, we force the 
last five steps to sample the correct value, and store the probability of the 
model sampling those values. We also evaluate the probability of reconstructing 
correctly the individual data points by not forcing the last five time steps to 
sample the correct value, but by storing the probability of reconstructing the 
correct value for each data point separately.
We run the MCMC chain 100 times and use the log of 
the mean of the likelihoods of predicting the correct value over these 100 runs.

When evaluating the performance of one-directional inference, we use a similar approach to MCMC. However, when evaluating the log-likelihood of the entire gap, we only construct it once in sequential order, and record the probabilities of reconstructing the correct value. When evaluating the probability of reconstructing the correct value for each data point separately, we use the same approach as for MCMC and sample the gap 100 times, recording for each step the probability of sampling the correct value. The result for each data point is the log of the mean of the likelihoods over these 100 runs.

On the Wikipedia data, we evaluate the GSN and NADE methods on 50~000 gaps on 
the test data. On the music data, all models are evaluated on all possible 
gaps of $g=5$ on the test data, excluding gaps that intersect with the first and 
last 10 time steps of a score. When evaluating the Bayesian MCMC with the unidirectional 
RNN, we have to significantly limit the size of the data set, as the method is 
highly computationally complex. We therefore run it on 1~000 gaps on the test 
data.

For NADE, we set the five time steps in the gap to the missing value token. We 
then reconstruct them one by one to the correct value, and record the 
probability of the correct reconstruction. We repeat this process for all 
possible permutations of the order in which to do the reconstruction, and 
therefore acquire the exact probability of the correct reconstruction given the 
model and the data. We also evaluate the individual character reconstruction 
probabilities by recording the probability of sampling the correct value given 
all other values in the gap are set to missing.

\subsection{Results}

\begin{table}[t]
\caption{
Negative Log Likelihood (NLL) for gaps of five time steps using different 
models (lower is better). In the experiments, GSN and NADE perform well, although they
are outperformed by Bayesian MCMC.}
\begin{tabular*}{\textwidth}{@{\extracolsep{\fill} } l c  c  c  c c  }
\hline\noalign{\smallskip}
\centering Inference strategy  & Wikipedia & Nottingham & Piano & Muse & JSB \\ 
\noalign{\smallskip}
\hline
\noalign{\smallskip}
GSN 		& 4.60 & 19.1 & \textbf{38.8} & 37.3 & \textbf{43.8} \\
NADE masked		& 4.86 & 19.0 & 40.4 & \textbf{36.5} & 44.3 \\
NADE 		& 4.88 & \textbf{18.5} & 39.4 & 34.7 & 44.6 \\
Bayesian MCMC 	& \textbf{4.37} & NA & NA & NA & NA  \\
One-way inference 	& 5.79 & 19.2 & 38.9 & 37.6 & 43.9  \\
\noalign{\smallskip}
\hline
\noalign{\smallskip}
One-gram & 23.3 & 145 & 138 & 147 & 118 \\
\hline
\end{tabular*}
\label{tab:bps}
\end{table}


From Table~\ref{tab:bps} we can see that the Bayesian MCMC 
method seems to yield the best results, while GSN or NADE outperform one-way inference. It is worth noting that in the most difficult data sets, piano and JSB, oneway inference performs very well. Qualitative 
examples of the reconstructions obtained with the GSN and NADE on the Wikipedia data are shown in 
Table~\ref{tab:samples} (supplementary material).

\begin{figure}[t]
    \centering
    \includegraphics[width=0.8\textwidth]{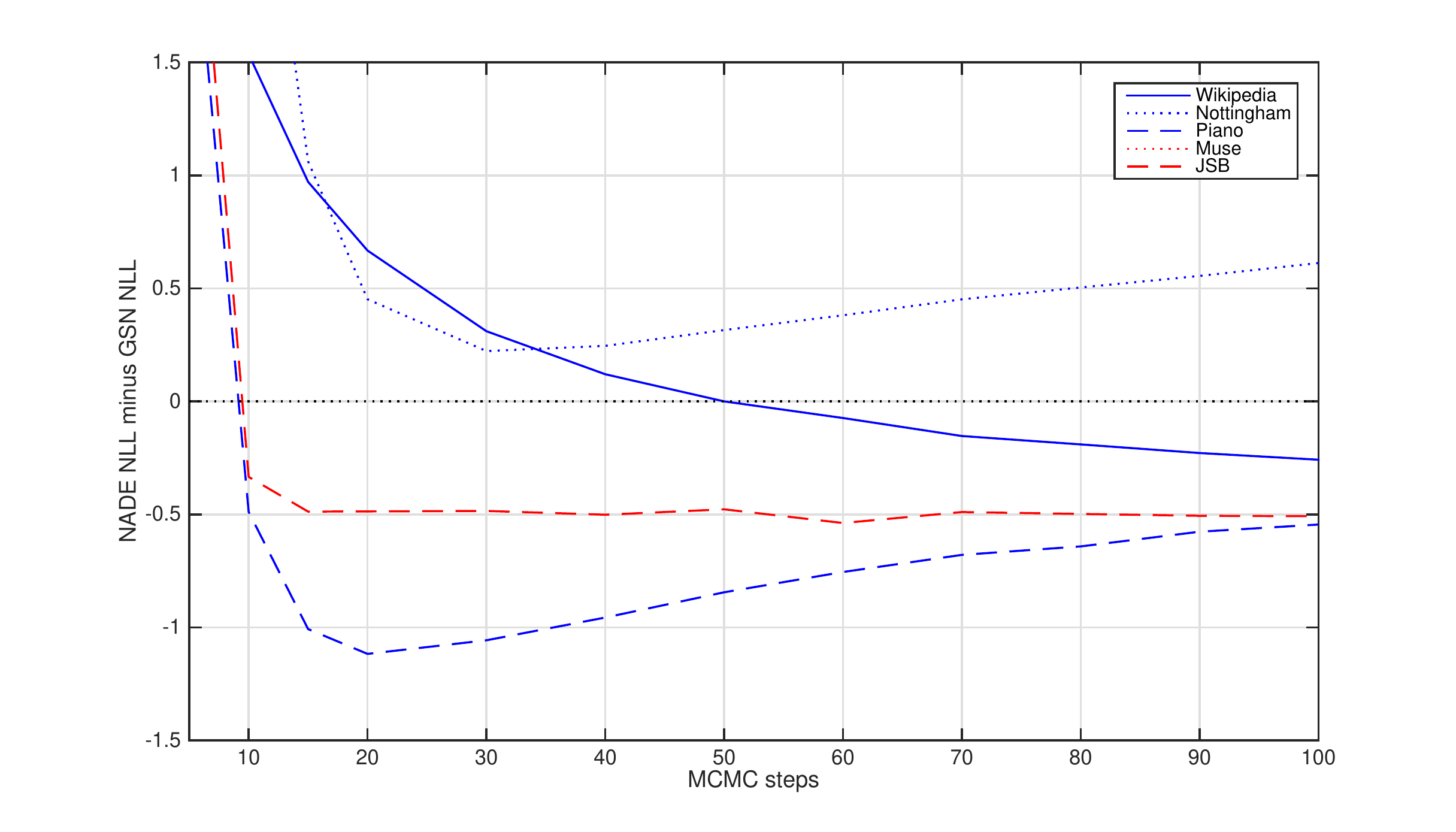}
    \caption{Difference in gap negative test log-likelihood between GSN and NADE for 
different data sets. We can see that GSN outperforms NADE after a certain 
threshold of MCMC steps. Note that the rising curves with music data sets might be explained
by the model being overfitted to the training set.}
    \label{fig:mcmc_iters}
    \vspace{-3mm}
\end{figure}

In order to get an indication of how the number of MCMC steps in the GSN approach affects 
performance, we plotted the difference in NLL of GSN and NADE of the test set as 
a function of the number of MCMC steps in Figure~\ref{fig:mcmc_iters} (supplementary material). The 
figure indicates that the music data sets mix fairly well, as the performance of 
GSN quickly saturates. However, for the Wikipedia data, the performance could 
probably be even further improved by letting the MCMC chain run for more than 
$M=100$ steps.

\begin{figure}[t]
    \centering
    \includegraphics[width=0.49\textwidth]{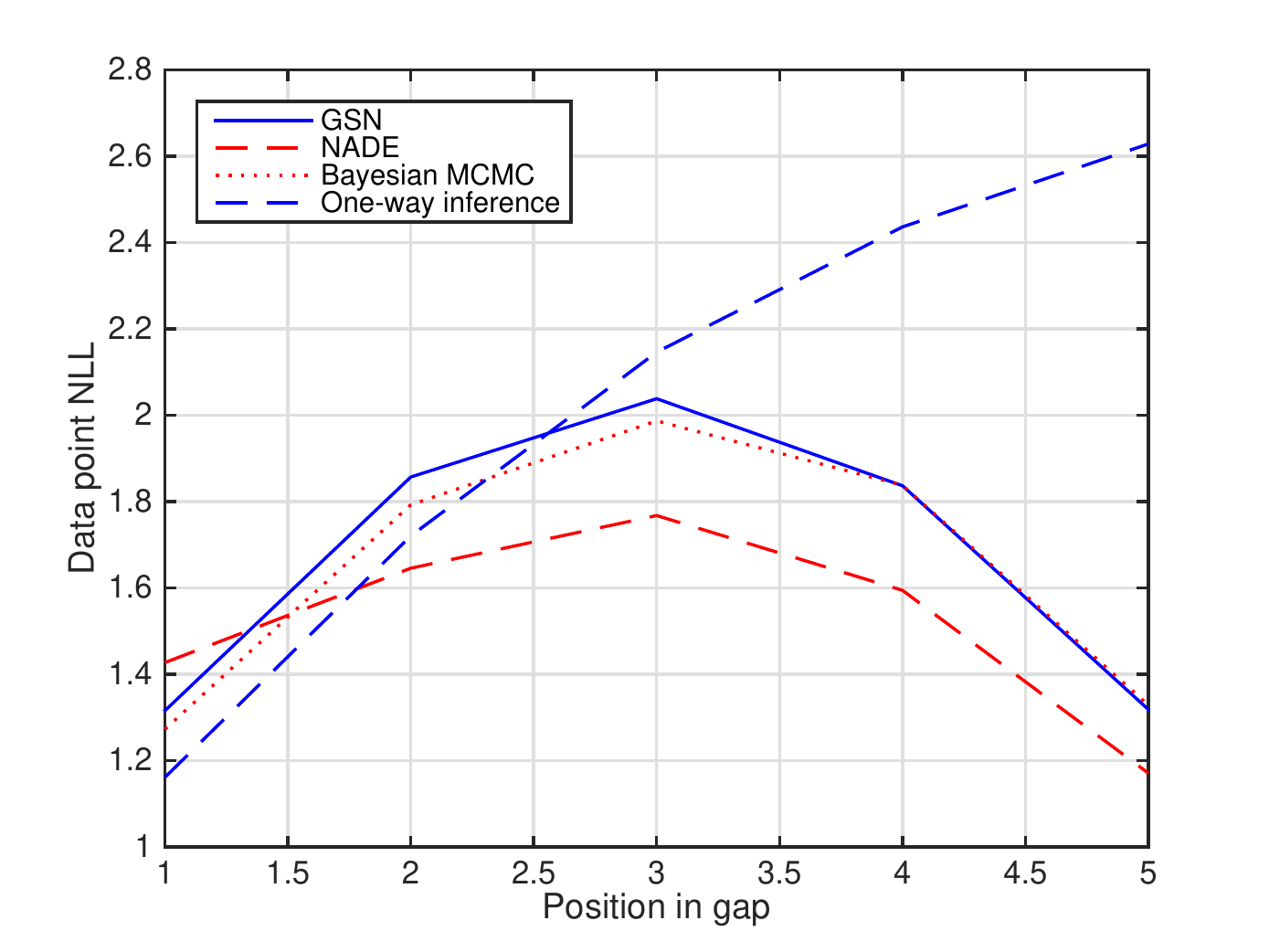}
    \includegraphics[width=0.49\textwidth]{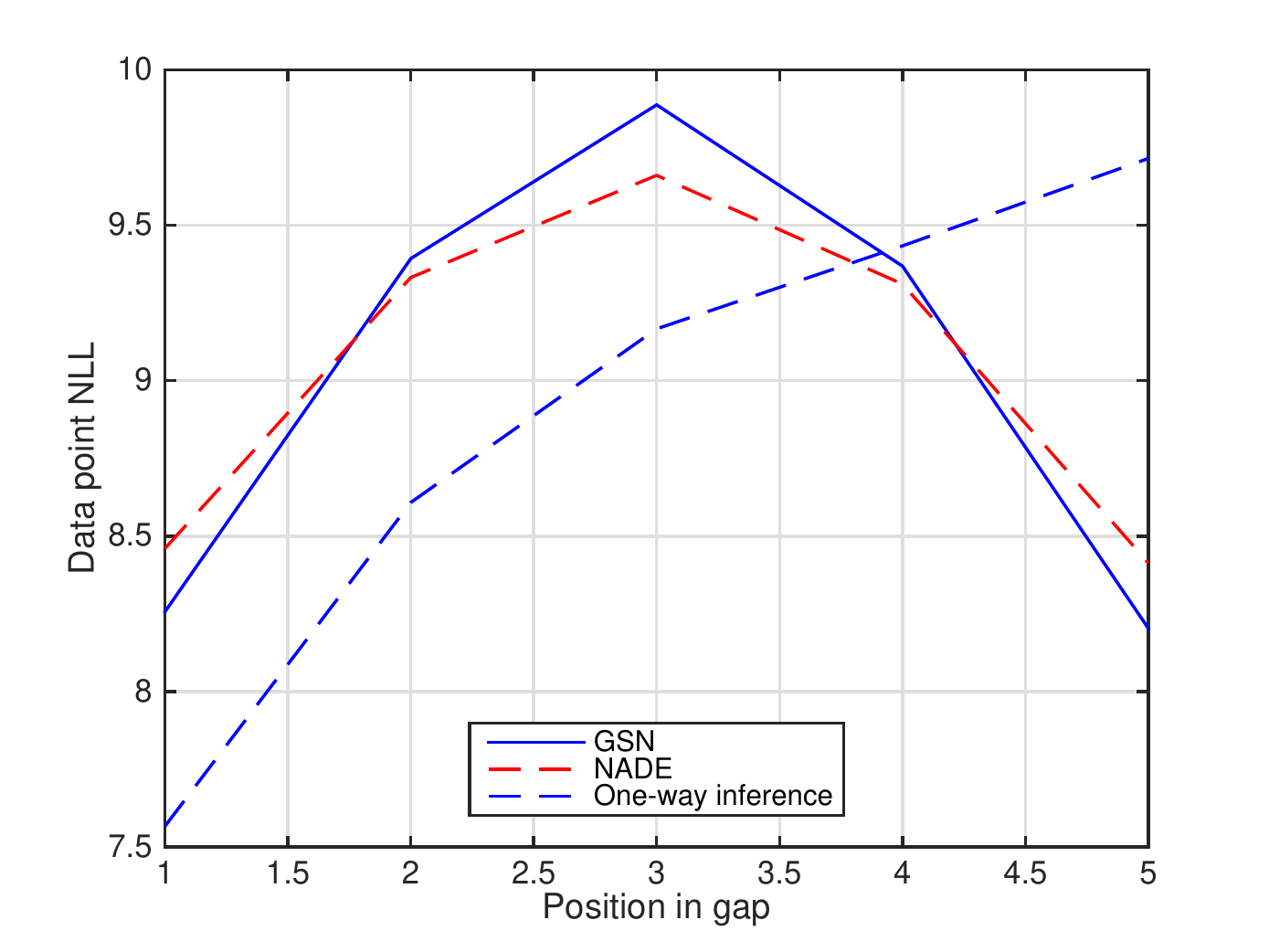}
    \caption{Average NLL per data point using different methods with the 
Wikipedia data set (left) and the Piano data set (right) for different positions 
in a gap of 5 consecutive missing values. The middle data point is the most difficult to estimate for the 
most methods, while the one-way inference cannot take future context into account 
making prediction of later positions difficult. For the left-most position in the gap, the one-way
inference performs the best since it does not require any approximations such as MCMC.}
    \label{fig:chars}
\end{figure}

In Figure~\ref{fig:chars} we have evaluated the NLL for the individual 
characters in the gaps of length five. As expected, all methods except for one-way inference are better at 
predicting characters close to both edges of the gap.

As a sanity check, we make sure our models have been successfully trained by evaluating 
the mean test log-likelihood of the BRNNs for gap sizes of one. In
Table~\ref{tab:nll_train} (supplementary material) we can see that the BRNNs expectedly outperform previously published
results with unidirectional RNNs, which indicates that the models have been trained successfully.

\section{Conclusion and Discussion}

Although recurrent neural networks have been used as generative models for time 
series data, it has not been trivial how to use them for inference in cases 
such as missing gaps in the sequential data.

In this paper, we proposed to use bidirectional RNNs as generative models for 
time series, with two probabilistic interpretations called GSN and NADE. Both 
provide efficient inference in both positive and negative directions in time, 
and both can be used in tasks where Bayesian inference of a unidirectional RNN 
is computationally infeasible.


The model we trained for NADE differed from the basic BRNN in several ways: 
Firstly, we artificially marked gaps of 5 consecutive points as missing, which 
should help in specializing the model for such reconstruction tasks. 
It would be interesting to study the effect of the missingness pattern used in training, on the learned representations and predictions. Secondly, in addition to using all outputs as the training signal, 
we tested using only the reconstructions of those missing values as the training signal. This reduces the effective amount of training that the 
model went through. Thirdly, the model had one more input (the missingness 
indicator) that makes the learning task more difficult.
We can see from Table~\ref{tab:nll_train} that the model we trained for NADE
where we only used the reconstructions as the training signal has a worse
performance than the BRNN for reconstructing single values. This indicates
that these differences in training have a significant impact on the quality
of the final trained probabilistic model.

We used the same number of parameters when training an RNN and a BRNN. The RNN can 
concentrate all the learning effort on forward prediction, and re-use the 
learned dependencies in backward inference by the computationally heavy 
Bayesian inference. It remains an open question which approach would work best 
given an optimal size of the hidden layers.

As future work, other model structures could be explored in this context, for 
instance the Long Short-Term Memory \citep{hochreiter1997long}. Specifically to 
our NADE approach, it might make sense to replace the regular additive 
connection from the missingness indicator input to the hidden activations in Eq.~(\ref{eq:bi_hf},\ref{eq:bi_hb}), by a 
multiplicative connection that somehow gates the dynamics mappings $\mathbf{W}_h^f$ and $\mathbf{W}_h^b$. Another direction to extend is to use a deep architecture with more hidden layers.

The midi music data is an example of a structured prediction task: Components 
of the output vector depend strongly on each other. However, our model assumes 
independent Bernoulli distributions for them. One way to take those dependencies 
into account is to use stochastic hidden units $\mathbf{h}_t^f$ and $\mathbf{h}_t^b$, 
which has been shown to improve performance on structured 
prediction tasks \citep{Raiko2015techniques}.
\citet{bayer2014learning} explored that approach, and reconstructed missing values in the middle of motion capture data. In their reconstruction method, the hidden stochastic variables are selected based on an auxiliary inference model, after which the missing values are reconstructed conditioned on the hidden stochastic variable values. Both steps are done with maximum a posteriori point selection instead of sampling. Further quantitative evaluation of the method would be an interesting point of comparison.


The proposed methods could be easily extended to continuous-valued data. As an 
example application, time-series reconstructions with a recurrent model has 
been shown to be effective in speech recognition especially under impulsive 
noise \citep{remes2011missing}.

\section*{Acknowledgements}

We thank KyungHyun Cho and Yoshua Bengio for useful discussions.
The software for the simulations for this paper was based on
Theano \citep{bastien-theano,bergstra-theano-scipy}.
Nokia has supported Mathias Berglund and the Academy of Finland has supported Tapani Raiko.

\small{
\bibliography{missingvalue}
\bibliographystyle{natbib}
}

\newpage

\begin{table}[t]
\caption{
Negative Log Likelihood (NLL) over the test data for the trained models. 
As a comparison, similar numbers are presented for the unidirectional RNNs trained 
by \citet{boulanger2012modeling} and by \citet{NIPS2013_5166}. The results serve as a 
sanity check that the trained models have been trained successfully. Note that 
the BRNN has more context than the unidirectional RNNs, and is hence expected 
to perform better measured by NLL. Also note that the training of the model for NADE without masking
is very similar to the training of the BRNN.}
\begin{tabular*}{\textwidth}{@{\extracolsep{\fill} } l c  c  c  c c  }
\hline\noalign{\smallskip}
\centering Inference strategy  & Wikipedia & Nottingham & Piano & Muse & JSB \\ 
\noalign{\smallskip}
\hline
\noalign{\smallskip}
NLL of BRNN & \textbf{0.37} & 3.23 & \textbf{6.82} & 5.68 & 7.80 \\
NLL of NADE masked & 0.55 & 3.32 & 7.42 & 6.48 & 8.51 \\
NLL of NADE & 0.41 & \textbf{2.89} & 7.05 & \textbf{5.54} & \textbf{7.59} \\
NLL of unidirectional RNN & 1.21 & 3.87 & 7.88 & 7.43 & 8.76 \\
\citet{boulanger2012modeling} & & 4.46 & 8.37 & 8.13 & 8.71 \\
\citet{NIPS2013_5166} & 1.12 &&&& \\
\hline
\end{tabular*}
\label{tab:nll_train}
\end{table}

\begin{table}[t]
\caption{
Random samples from reconstructed gaps (underlined) using either NADE (left) or 
GSN (right). Note that double spaces are difficult to distinguish from single 
spaces.
}
\centering
\begin{tabular}{ l  l }
\hline
\noalign{\smallskip}
NADE  & GSN \\
\noalign{\smallskip}
\hline
\noalign{\smallskip}
\begin{bookman} s practice their sho\underline{rt fo}r as long as possibl \end{bookman}  & \begin{bookman} s practice their sho\underline{w,d a}r as long as possibl \end{bookman}  \\
\begin{bookman} nd nephews through c\underline{opien}e. It was reported o \end{bookman}  & \begin{bookman} nd nephews through c\underline{lubs.}e. It was reported o \end{bookman}  \\
\begin{bookman} but originally a nuc\underline{lear }bunker to protect th \end{bookman}  & \begin{bookman} but originally a nuc\underline{lear }bunker to protect th \end{bookman}  \\
\begin{bookman}  the Opera" and "The\underline{ Heas}", have been fully r \end{bookman}  & \begin{bookman}  the Opera" and "The\underline{reove}", have been fully r \end{bookman}  \\
\begin{bookman} e III. Dunmore died \underline{in Ma}rch 1809 and was suc \end{bookman}  & \begin{bookman} e III. Dunmore died \underline{in Ma}rch 1809 and was suc \end{bookman}  \\
\begin{bookman} ch fades from golden\underline{ aill}ow at the center thr \end{bookman}  & \begin{bookman} ch fades from golden\underline{ly sh}ow at the center thr \end{bookman}  \\
\begin{bookman} Colorado state champ\underline{ion w}ho is credited with  \end{bookman}  & \begin{bookman} Colorado state champ\underline{ions }ho is credited with  \end{bookman}  \\
\begin{bookman} ing Bushroot, Liquid\underline{a. HL} and Megavolt).  His \end{bookman}  & \begin{bookman} ing Bushroot, Liquid\underline{lands} and Megavolt).  His \end{bookman}  \\
\begin{bookman} acial lake bed known\underline{ as t}he Black Dirt Region \end{bookman}  & \begin{bookman} acial lake bed known\underline{ of t}he Black Dirt Region \end{bookman}  \\
\begin{bookman} e all ancient leader\underline{sidre}d it to laud their a \end{bookman}  & \begin{bookman} e all ancient leader\underline{ cann}d it to laud their a \end{bookman}  \\
\begin{bookman} ted November 2005. T\underline{hey s}eries also featured  \end{bookman}  & \begin{bookman} ted November 2005. T\underline{he  s}eries also featured  \end{bookman}  \\
\begin{bookman} TR amyloid is extrac\underline{tedli}ar.  Treatment of TT \end{bookman}  & \begin{bookman} TR amyloid is extrac\underline{t s w}ar.  Treatment of TT \end{bookman}  \\
\begin{bookman} hile the gigantic "S\underline{ton s}aurus sikanniensis", \end{bookman}  & \begin{bookman} hile the gigantic "S\underline{"So S}aurus sikanniensis", \end{bookman}  \\
\begin{bookman}  area. Initially one\underline{ othe}r compartment was an \end{bookman}  & \begin{bookman}  area. Initially one\underline{ othe}r compartment was an \end{bookman}  \\
\noalign{\smallskip}
\hline
\end{tabular}
\label{tab:samples}
\end{table}

\end{document}